\documentclass{article}

\PassOptionsToPackage{numbers, compress}{natbib}
\usepackage[preprint]{neurips_2020}

\usepackage[utf8]{inputenc} % allow utf-8 input
\usepackage[T1]{fontenc}    % use 8-bit T1 fonts
\usepackage{hyperref}       % hyperlinks
\usepackage{url}            % simple URL typesetting
\usepackage{booktabs}       % professional-quality tables
\usepackage{amsfonts}       % blackboard math symbols
\usepackage{nicefrac}       % compact symbols for 1/2, etc.
\usepackage{microtype}      % microtypography
\usepackage{color}
\usepackage{amsmath}
\usepackage{mmstyle}
\usepackage{multirow}
\usepackage{booktabs}
\usepackage{wrapfig}
\usepackage{graphicx}
\usepackage{caption}
\usepackage{subfig}

\title{Unsupervised Landmark Learning \\ from Unpaired Data}
\author{
    Yinghao Xu $^{\dag}$ \quad Ceyuan Yang $^{\dag}$ \quad Ziwei Liu \quad  Bo Dai \quad Bolei Zhou \\
    Department of Information Engineering\\
  The Chinese University of Hong Kong \\
  \texttt{$\{$xy119,yc019,zwliu,bdai,bzhou$\}$@ie.cuhk.edu.hk} 
}

\begin{document}

\maketitle

\begin{abstract}
Recent attempts for unsupervised landmark learning leverage synthesized image pairs that are similar in appearance but different in poses.
These methods learn landmarks by encouraging the consistency between the original images and the images reconstructed from swapped appearances and poses.
While synthesized image pairs are created by applying pre-defined transformations, they can not fully reflect the real variances in both appearances and poses. 
In this paper, we aim to open the possibility of learning landmarks on unpaired data (i.e. unaligned image pairs) sampled from a natural image collection, so that they can be different in both appearances and poses.
To this end, we propose a cross-image cycle consistency framework ($C^3$) which applies the swapping-reconstruction strategy twice to obtain the final supervision.
Moreover, a cross-image flow module is further introduced to impose the equivariance between estimated landmarks across images.
Through comprehensive experiments, our proposed framework is shown to outperform strong baselines by a large margin.
Besides quantitative results, we also provide visualization and interpretation on our learned models, which not only verifies the effectiveness of the learned landmarks, but also leads to important insights that are beneficial for future research.\footnote{Code and models are available at \href{https://github.com/justimyhxu/ULTRA}{this link.} \\ \indent $\dag$ indicates equal contribution.}

\end{abstract}

%% Intro
\section{Introduction}\label{sec:intro}

Landmark, a robust and consistent representation for object structures, has been widely adopted in object-centric visual understanding, such as recognizing actions \cite{yan2018spatial,du2015hierarchical}, analyzing human behaviors \cite{hourglass,openpose}, and understanding video dynamics \cite{video_dynamic, video_inter}. Early attempts of landmark detection treat landmarks as human-defined keypoints and resort to a large amount of manual annotations to supervise the model learning.
However, manual annotations are usually costly to obtain. 
Meanwhile, human defined keypoints are likely to be inconsistent with the ideal semantics of landmarks.

To avoid aforementioned issues, unsupervised landmark learning \cite{thewlis2017unsupervised,thewlis2019unsupervised,abiv,jakab2018unsupervised,Zhang_2018_CVPR,denton2017unsupervised} has gained substantial attention.
Compared to supervised landmark learning that relies on human-defined keypoints, unsupervised landmark learning regards landmarks as pose-specific semantic points that facilitate the reconstruction of the original image.
Such an assumption has successfully led to geometric and semantic consistent landmarks learned in a data-driven manner.
Specifically, existing methods often adopt a \emph{swapping-reconstruction} strategy.
Starting from a pair of two images that are the same in appearance but different in poses, they disentangle two images' appearances and poses, respectively in the form of feature vectors and landmarks, and reconstruct one image using its pose and the appearance of its counterpart.
Since it is hard to collect pairs of images in reality that are the same in appearance but different in poses, synthesizing such pairs is further performed by taking a reference image and applying pre-defined transformations (\eg~affine and TPS transformations) on it to obtain its counterpart with a different pose.

\begin{figure}[t]
    \centering
    \includegraphics[width=0.95\textwidth]{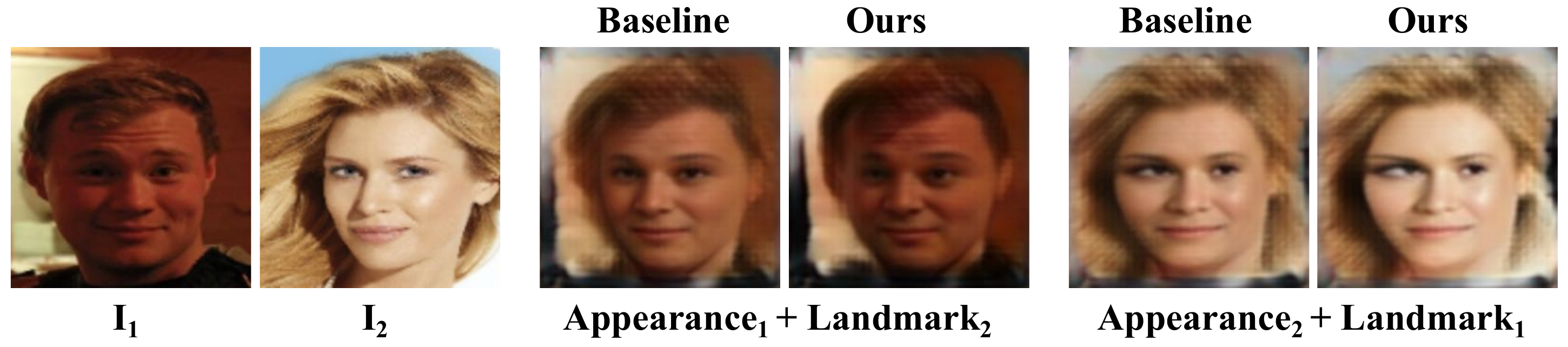}
%    \vspace{-5pt}
    \caption{
        \textbf{Limitations of existing methods.} The extracted pose-specific representation may contain some appearance-specific variation factor (\eg skin color).
        Our approach shows more disentangled synthesized results
}
\label{fig:illumination}
  \vspace{-1.5em}
\end{figure}

Though effective, these methods are still limited in several aspects:
Firstly, synthesizing image pairs via pre-defined transformations can hardly cover all cases encountered in real scenarios.
Therefore, the learned landmarks may not generalize well.
Moreover, even when using data collected from real scenarios,
learning only on image pairs that are the same in appearance but different in poses can result in an incomplete disentanglement of appearance and pose. Certain appearance-specific varying factors (\eg~illumination) are involved in pose-specific landmark representations, since during training models have never seen image pairs with varying appearances.
Consequently, when transforming a target image using its appearance and pose extracted from a reference image,
the transformed image may contain some appearance-specific semantics of the reference image, as shown in Figure \ref{fig:illumination}.

To enable learning landmarks from unpaired images that not only varying in poses but also varying in appearances, we propose a cross-image cycle consistency ($C^3$) framework in this paper.
The key insight is to apply \emph{swapping-reconstruction} strategy twice to obtain two reconstructed images that are supposed to exactly resemble the original two images respectively.
In this way, the learned models are better aware of appearance-specific variance as well as pose-specific variance.
The learned landmarks can thus better meet its original definition as pose-specific semantic points.
To avoid trivial solution of such a cycle consistent approach, we further regularize the learning process by considering the equivariance property of landmarks.
Specifically, we estimate a transformation map between two images via a cross-image flow module and constrain the detected landmarks to be equivariant with respect to given transformation map.
Through a comprehensive empirical study, landmarks learned by our framework are shown to outperform previous methods by a large margin.
Notably, our framework can also lead to an improvement of semantic alignment \cite{RTN,weakalign,cnngeo,Neighbourhood}, a closely related task to landmark learning, by simply adapting the cross-image flow module.

%% Related work
\section{Related Work}\label{related}

\noindent\textbf{Unsupervised Landmark Detection.}
Previous attempts \cite{TCDCN,MTCNN,cascadedCNN} for object landmark detection often adopt a supervised setting,
where a set of images with human provided keypoints is used 
to minimize the distances between model predicted points and human provided ones.
Recently, researchers resort to unsupervised landmark detection \cite{landmarkfactorize,thewlis2017unsupervised}
due to its superior performance compared to supervised landmark detection.
Instead of utilizing human annotations,
most of these methods for unsupervised landmark detection learn models 
with synthesized image pairs that are same in appearance but different in poses,
which are created by transforming one natural image using pre-defined transformations such as affine and TPS transformations.
Specifically, \cite{Zhang_2018_CVPR,jakab2018unsupervised} build conditional-encoder upon the traditional auto-encoder network
to discover unsupervised landmark representations with meaningful visual semantics.
Restricted Boltzmann Machines \cite{hintonbz} and InfoGAN \cite{infogan} are further introduced to impose a certain structure in the latent space.
The core idea of \cite{thewlis2017unsupervised} is also shared by \cite{abiv,lorenz2019unsupervised},
which factorize image semantics into background semantics and object-related semantics,
and learn to detect landmarks by disentangling different semantics and reconstructing original images.
Compared to previous methods that rely on synthesized image pairs, 
in this work, we enable the possibility of unsupervised landmark learning with unaligned image pairs sampled from natural images,
which can be different in both appearances and poses.
We achieve this by introducing a cross-image cycle consistent framework and an additional cross-image flow module,
which implicitly disentangle the appearances and poses of images of objects 
and regularize the equivariance of them across unaligned image pairs.

\noindent\textbf{Landmark and Appearance Disentanglement.}
Our work is also related to landmark and appearance disentanglement,
which is a popular topic in tasks of representation learning.
Specifically, constraining generative models on the appearance information to disentangle different semantics has been introduced in \cite{denton2017unsupervised}.
Moreover, \cite{desjardins2012disentangling} introduce a holistic model for disentanglement learning.
While our proposed framework can learn to disentangle pose and appearance in an unsupervised way,
we use landmarks as the pose representation and can disentangle appearance and pose from images of high-dimensional data such as faces.
\section{Methodology}\label{sec:method}

\begin{figure}
    \centering
    \includegraphics[width=1\textwidth]{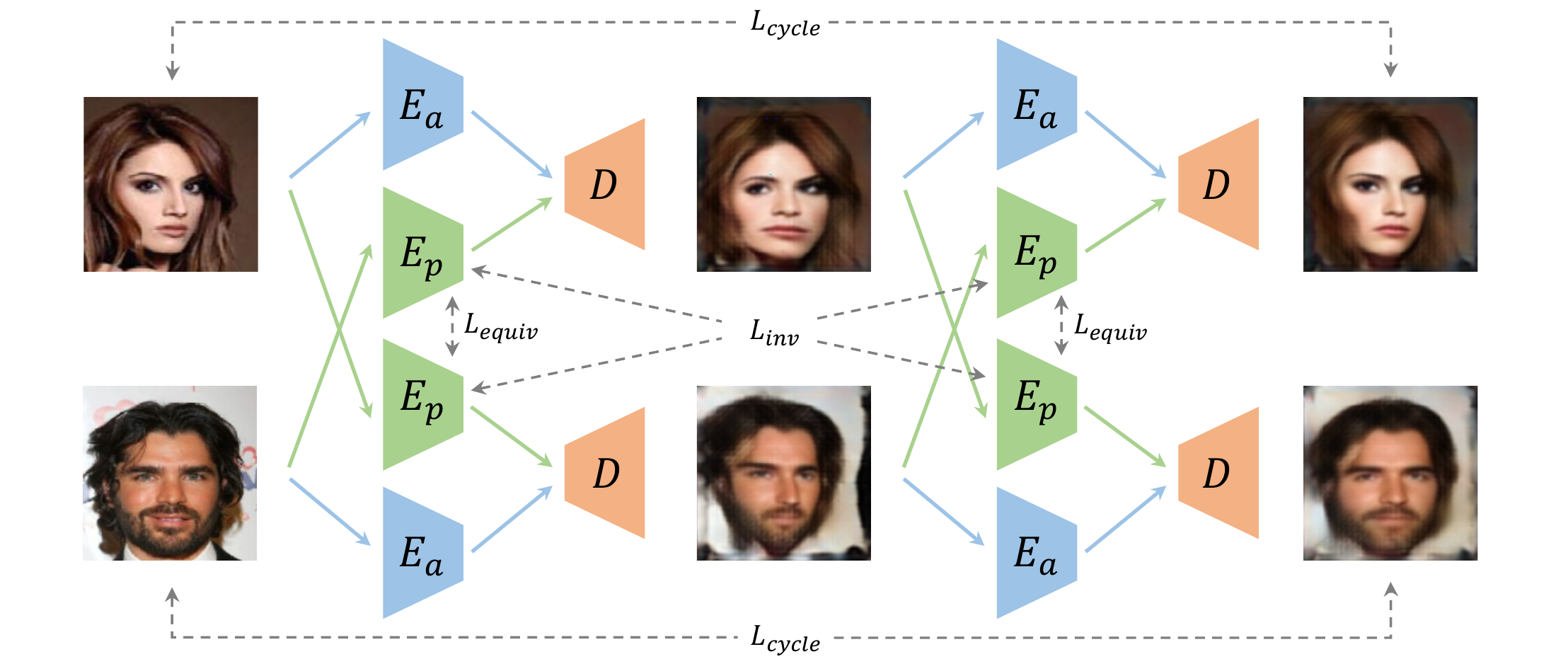}
    \caption{
        \textbf{Framework of $C^3$.} To enable the learning from unpaired data, our method conducts the \emph{swapping-reconstruction} strategy twice 
        and encourages the reconstruted images to resemble the original. 
        A cross-image flow module further regularizes the equivariance of detected landmarks
}
    \label{fig::framework}
    \vspace{-1.5em}
\end{figure} 

While supervised landmark learning encourages models to resemble human defined keypoints,
it remains a question that whether human defined keypoints are consistent with the ideal semantics of landmarks.
Alternatively, recent approaches belonging to unsupervised landmark learning 
regard pose-specific semantic points
that dominate the reconstruction of an image,
and learn them from synthesized image pairs 
where one image is transformed from the other one by pre-defined transformations (\eg~affine and TPS transformations)
so that they are same in appearance but different in poses.
However, since synthesized pairs contain only images with the same appearance 
they can not well reflect the variance of appearance.
Consequently the aforementioned unsupervised landmark learning methods 
may mistakenly include appearance-specific semantics in the learned landmarks as shown in Figure \ref{fig:illumination},
leading to unsatisfactory landmarks.

To enable unsupervised landmark learning directly on unpaired natural images with varying appearances and poses,
in this paper we propose a cross-image cycle consistent framework ($C^3$),
together with a cross-image flow module to further regularize the learning process.
Compared to previous methods,
landmarks learned by the proposed framework are shown to be more consistent across different images,
and better align with their roles as pose-specific semantic points. 
In the following, we will respectively introduce details of the cross-image cycle consistent framework
and the cross-image flow module in Sec.\ref{subsec:framework} and Sec.\ref{subsec:flow}.

\subsection{Landmark Learning via Cross-Image Cycle Consistency Framework}\label{subsec:framework}
Following previous approaches, we regard landmarks as a kind of pose-specific semantic representation,
and refer to pose of an image as the set of $K$ landmarks $L = \{l_1, ..., l_k\}$ detected from it.
We aim at learning landmarks using not only synthesized image pairs that are same in appearance and different in poses,
but also natural image pairs with both varying appearances and poses.
The key is to reconstruct images with the appearance and pose originated from different images 
and establish various consistencies among these reconstructed images.

An overview of our framework is included in Figure \ref{fig::framework}.
Specifically, given a set of images $\cD = \{I_1, ..., I_m\}$, 
our main framework consists of a backbone network $N$, a pose extractor $E_p$, an appearance extractor $E_a$, and an image decoder $D$.
For any image $I$,
we can extract its raw feature map $\vf$ via the backbone network $N$.
On top of $\vf$, we can then obtain the appearance and pose of image $I$ with two extractors $E_a$ and $E_p$,
which can be further used by the image decoder to reconstruct the original image: 
\begin{align}
    \va & = E_a(\vf), \qquad \vp = E_p(\vf), \notag \\
    I & = D(\va, \vp),\notag
\end{align}
where $\va$ and $\vp$ are both feature maps, and the number of channels of $\vp$ equals to $K$. 
It's worth mentioning, in order to bridge the output of $E_p$ and the landmarks $L$ we expect to acquire,
in practice we follow \cite{DRIT} to transform $\vp$ by a two-step process.
At the first step, we normalize each channel of raw feature map $\vp$ using a softmax operation
and treat it as the probability map of a landmark.
We thus get the estimated coordinate $(x_t, y_t)$ of $t$-th landmark by taking expectations over $t$-th channel of $\vp$.
At the second step, we replace $t$-th channel of $\vp$ by a gaussian noise map whose mean is located at $(x_t, y_t)$.
After this process, the transformed $\vp$ can then be regarded as 
a differentiable representation of landmarks and a feature map for succeeding operations.

For a pair of images $(I_i, I_j)$,
we can not only reconstruct the original images as mentioned above,
but also follow a \emph{swapping-reconstruction} strategy and reconstruct two swapped images 
each of which contain the appearance of one image and the pose of the other:
\begin{align}
    \va_i & = E_a(N(I_i)), \qquad \vp_i = E_p(N(I_i)), \notag \\
    \va_j & = E_a(N(I_j)), \qquad \vp_j = E_p(N(I_j)), \notag \\
    I_{i, j} & = D(\va_i, \vp_j), \notag \\
    I_{j, i} & = D(\va_j, \vp_i). \notag
\end{align}
To learn $N$, $E_p$, $E_a$ and $D$ with any natural image pair $(I_i, I_j)$ sampled from $\cD$,
we propose the cross-image cycle consistency framework 
which conducts the swapping-reconstruction strategy \emph{twice} as
\begin{align}
    \va_i^\prime & = E_a(N(I_{i, j})), \qquad \vp_j^\prime = E_p(N(I_{i, j})), \notag \\
    \va_j^\prime & = E_a(N(I_{j, i})), \qquad \vp_i^\prime = E_p(N(I_{j, i})), \notag \\
    I_i^\prime & = D(\va_i^\prime, \vp_i^\prime), \notag \\
    I_j^\prime & = D(\va_j^\prime, \vp_j^\prime).\notag
\end{align}  
Consequently, we can encourage $I_i^\prime$ and $I_j^\prime$ to resemble the real images $I_i$ and $I_j$ by applying a perceptual loss \cite{perceptual}:
\begin{align}
    \cL_\mathrm{cycle} = \cP(I_i, I_i^\prime) + \cP(I_j, I_j^\prime),
\end{align}
where $\cP$ is implemented with a VGG network as in \cite{vgg}.
Besides, we can also add additional constraints on the invariance of extracted poses:
\begin{align}
    \cL_\mathrm{inv} = \Vert \vp_i^\prime - \vp_i \Vert_2 + \Vert \vp_j^\prime - \vp_j \Vert_2.
\end{align}

\subsection{Regularization via Cross-Image Flow Module}\label{subsec:flow}

While the proposed $C^3$ can help learn landmarks using natural image pairs with varying appearances and poses,
to further constrain the correspondences of learned landmarks across different images,
we propose a learnable cross-image flow module to estimate the transformation map between two images,
and thereafter apply a correspondence constraint on top of the estimated map.

Specifically, a transformation map $T^{i\rightarrow j}$ establishes the location correspondences when warping image $i$ to image $j$
or their feature maps at the same level.
To estimate $T^{i\rightarrow j}$ and $T^{j\rightarrow i}$ for any pair of images $(I_i, I_j)$ sampled from $\cD$,
we at first reuse the backbone network $N$ to extract two feature maps $\vf^\prime_i$ and $\vf^\prime_j$.
A 4D tensor $\mC$ containing the element-wise cosine similarity between $\vf^\prime_i$ and $\vf^\prime_j$ is then computed.
\ie~The value of $\mC$'s entry $(x_i, y_i, x_j, y_j)$ is 
the cosine similarity between the feature vector at $(x_i, y_i)$ of $\vf^\prime_i$
and the feature vector at $(x_j, y_j)$ of $\vf^\prime_j$.
We can then apply a convolutional layer with a 4D kernel to further process $\mC$.
In practice, since a convolutional layer with a 4D kernel are computationally expensive,
we replace it with two separate 2D convolutions:
\begin{align}
    \hat{\mC} = \mW_1 \otimes (\mW_2 \otimes \mC),
\end{align} 
where $\otimes$ denotes the convolution operator, and $\mW$ is a convolution kernel.
Estimation of $T^{i \rightarrow j}$ and $T^{j \rightarrow i}$ based on $\hat{\mC}$ follows:
\begin{align}
            S^{i\rightarrow j}(x_j, y_j) & = \mathrm{softmax}(\hat{\mC}(*, *, x_j, y_j)), \notag\\  
    T^{i\rightarrow j}(x_j, y_j) & = \argmax_{(x_i, y_i)} S^{i \rightarrow j}(x_j, y_j), \notag\\
            S^{j\rightarrow i}(x_i, y_i) & = \mathrm{softmax}(\hat{\mC}(x_i, y_i, *, *)), \notag\\
    T^{j\rightarrow i}(x_i, y_i) & = \argmax_{(x_j, y_j)} S^{j \rightarrow i}(x_i, x_i), \notag
\end{align}
where $S^{i\rightarrow j}$ and $S^{j \rightarrow i}$ are respectively score maps of location correspondences for two images.

Indeed, transformation maps $T^{i \rightarrow j}$ and $T^{j \rightarrow i}$ can 
reflect the semantic correlations between landmarks of two images.
We thus apply an equivariance loss: 
\begin{align}
    \cL_\mathrm{equiv} = \Vert \vp_i - T^{j \rightarrow i} \circ \vp_j \Vert_2 + 
 \Vert \vp_j - T^{i \rightarrow j} \circ \vp_i \Vert_2,
\end{align}
where $\circ$ denotes the warping operation.

\paragraph{Final Loss Function.}
The final loss function consists of all losses mentioned above:
\begin{align}
    \cL_\mathrm{total} = \lambda_\mathrm{cycle} \cL_\mathrm{cycle} + \lambda_\mathrm{equiv}\cL_\mathrm{equiv} + \lambda_\mathrm{inv} \cL_\mathrm{inv},
\end{align}
where $\{\lambda_\mathrm{cycle}, \lambda_\mathrm{equiv}, \lambda_\mathrm{inv}\}$ are balancing coefficients.
It worth noting that this loss function involves no prior on desired landmarks,
and puts no restrictions on the used image pairs.

%% Experiment
\section{Experiments}\label{sec:exp}

To evaluate the proposed method thoroughly, we first conduct a quantitative comparison with other state-of-the-art approaches on facial landmark detection in Sec.\ref{exp:landmark}.
We also include an ablation study and an interpretation to analyze components of our framework and the disentangled representations emerging from our framework.

In addition, in Sec.\ref{exp:se_align} we verify the effectiveness of our proposed cross-image flow module on semantic alignment, a task closely related to landmark detection that emphasizes the consistency of learned landmarks across different images.

\subsection{Implementation Details}\label{subsec:imp}
\textbf{Architectures.} 
ResNet-50 \cite{resnet} pretrained on ImageNet \cite{imagenet} is served as our backbone network $N$.
We use the outputs of the last two residual blocks respectively as the features $\vf^\prime$ and $\vf$ used in Sec.\ref{subsec:framework} and Sec.\ref{subsec:flow}.
The appearance extractor $E_a$ consists of two stacked $3\times3$ convolutional layers with ReLU \cite{relu}, and its output is a feature map contains 256 channels.
A similar structure is chosen for $E_p$ with an additional $1\times1$ convolutional layer to produce a differentiable landmark representation that contains $K$ channels as described in Sec.\ref{subsec:framework}.
The decoder $D$ takes the concatenation of appearance and pose feature maps as the input to decode an image.
More details about the architectures of our framework are available in the supplementary materials.

\textbf{Datasets.} 
For landmark detection, CelebA \cite{celeba}  is used for training, which contains around 20k face images from 10k identities. 
We use MAFL \cite{MTCNN} and AFLW \cite{AFLW} for testing, % with a tight bounding box around face and 5 facial landmarks.
All images are resized to $128 \times 128$.
To ensure a fair comparison, identities appeared in CelebA are removed from MAFL and AFLW.
As for semantic alignment, we evaluate the cross-image flow module on PF-Pascal \cite{ham2017proposal},
which includes 1,351 semantically related image pairs belonging to 20 object categories for the PASCAL VOC dataset \cite{pascal}.

\subsection{Facial Landmark Detection}\label{exp:landmark}
 
\textbf{Setup.}
Adam \cite{adam} is adopted as our optimizer.
We train our model on $4$ GPUs with a total batch size of 200. 
The initial learning rate is 0.0005 and will be decayed with a factor of 10 at $100$ and $150$ epochs (200 epochs in total). 
To prevent a trivial solution, we pretrain several parts independently using synthesized pairs as the initialization.
More details are presented in the supplementary materials.

\begin{table}[h]
\centering
\captionsetup[subfloat]{captionskip=4pt, position=bottom}
    \caption{
        \textbf{Comparison with other methods.} We report MSE metric normalized by IOD (inter-ocular distance) on MAFL \cite{MTCNN} and AFLW \cite{AFLW}.
        $K$ denotes the number of landmarks detected
        }
    \label{tab:main_results_ld}
    
    \subfloat[Comparison with the state-of-the-art \emph{Unsupervised / Self-Supervised} landmark detection methods  ]{
        \begin{tabular}{lccc}
            \toprule
            Method     & $K$     & MAFL    & AFLW  \\
            % \midrule \multicolumn{4}{c} { \emph{Unsupervised / Self-Supervised} } \\
            \midrule
            Fact \cite{landmarkfactorize}             & 10/30  &  NA/7.15    & NA/NA \\
            Struc.\cite{Zhang_2018_CVPR}               & 10/30  & 3.46/3.16   & 7.01/6.58 \\
            Cond. \cite{jakab2018unsupervised}       & 10/30  & 3.19/2.58   & 6.86/6.31 \\
            Joint. \cite{jeon2019joint}                      & 10/30  & 3.33/2.98   & 7.17/6.51 \\
            Shape. \cite{lorenz2019unsupervised}                         & 10/30  & 3.24/NA     & NA/NA \\
            \midrule
            Ours                                           & 10/30 & \textbf{3.11}/ \textbf{2.54} & \textbf{6.75}/\textbf{6.26}  \\
            \bottomrule
            \end{tabular}
    }\hspace{10pt}
    \subfloat[Comparison with \emph{Supervised} landmark detection methods]{
        \begin{tabular}{lccc}
            \toprule
            Method        & MAFL    & AFLW  \\
            % \midrule \multicolumn{3}{c} { \emph{Supervised } } \\
            \midrule
                RCPR \cite{RCPR}                               & -       & 11.60 \\
                CFAN \cite{CFAN}                                & 15.84   & 10.94 \\
                Cas CNN \cite{cascadedCNN}                 & 9.73    & 8.97 \\         
                TCDCN \cite{TCDCN}                            & 7.95    & 7.65 \\
                MTCNN \cite{MTCNN}                             & 5.39    & 6.90  \\
            \midrule
            Ours                                       & \textbf{2.54} & \textbf{6.26}  \\
            \bottomrule
            \end{tabular}
    }
    \vspace{-20pt}
\end{table}

\textbf{Quantitative Evaluation.}
Following the standard quantitative protocol for evaluating unsupervised landmark detection,
when testing we train a linear regression model to slightly adjust the position of detected landmarks
and fit the manually-annotated ground-truths in a supervised manner.
In other words, we train our framework on CelebA in an unsupervised manner and evaluate the detected landmarks on MAFL \cite{MTCNN} and AFLW \cite{AFLW} via supervised regressions.
MSE normalized by the inter-ocular distance expressed in the form of a percentage is reported as the metric,
following \cite{TCDCN}.

Table \ref{tab:main_results_ld} shows the results of ours and previous approaches in terms of both 10 and 30 landmark points.
We can see that compared to supervised learning,
methods that detect landmarks in an unsupervised manner obtain significantly better results, 
which indicates that human provided keypoints are limited in terms of their cross-image consistency.
On the other hand, among all the methods based on unsupervised landmark learning,
our method outperforms all the baselines by a clear margin across datasets and different numbers of landmark points.
Such results have verified our hypothesis that learning via synthesized image pairs following pre-defined transformations may lead to an incomplete disentanglement of appearance and pose, and further degrade the performance.
In contrast, our method enables the possibility to learn directly on natural image pairs with varying appearances and poses, 
and the disentanglement is thus better conducted,
which can further be verified in Figure \ref{fig:illumination}.

\begin{wraptable}{r}{4cm}
\vspace{-18pt}
    \begin{center}
    \caption{Ablation study on $C^3$ and flow module}
    \label{tab:ablationld}
    % \begin{center}
    \begin{tabular}{cc|c}

        \toprule
        Cycle   &  Flow  & MSE \\
        \midrule
                            &                 &  3.31  \\
        \checkmark          &                 &  3.18  \\
                            &   \checkmark    &  3.30  \\
        \checkmark          &   \checkmark    &  \textbf{3.11}  \\
        \bottomrule
        \end{tabular}
    \end{center}
    %\end{table}
    \vspace{-15pt}
    \end{wraptable}

\textbf{Ablation Study.} 
In order to validate two key components of our method (\ie $C^3$ and cross-image flow module), we conduct an ablation study on MAFL \cite{MTCNN}. 
The results are included in Table \ref{tab:ablationld}.
Specifically, when removing the $C^3$, our models can only learn from synthesized image pairs as in previous methods.
Consequently, semantic flow module can hardly help the landmark learning due to the limited variance of synthesized image pairs.
However, when learning directly from natural image pairs \ie adopting $C^3$, 
the cross-image flow module can make the landmarks more consistent across different images.
Finally, the proposed $C^3$ can leverage both appearance and landmark variances of natural images to improve landmark detection,
even without the flow module.

\subsection{Visualization and Interpretation} \label{exp:disentangle}
\noindent{\textbf{Cross-Image Consistency of Detected Landmarks.}}
An important aspect of unsupervised landmark detection is that the detected landmarks should be consistent across different images.
For example, despite of the difference in facial orientations and person identities,
the point near the nose should at the same location when transformed from one image to another.
While the standard metric focuses on the similarity between detected landmarks and manually-annotated ones,
here we evaluate the consistency of detected landmarks across different images,
where we use the transformation map estimated by our cross-image flow module to do the warping.

\begin{figure}[t]
    \centering
    \includegraphics[width=1\textwidth]{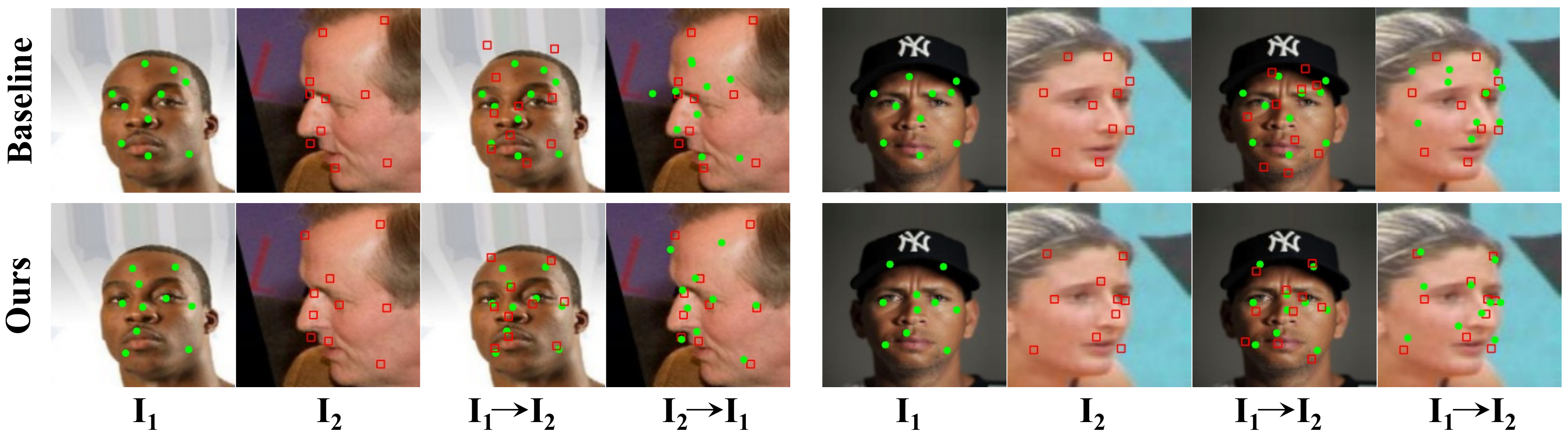}
    \caption{
        \textbf{Cross-Image Landmark Consistency.} 
        Learning from different instances (Ours) allows to obtain more consistent landmarks than those from the same (Baseline), especially when the orientation and gender are significantly different. 
        %Landmarks are wrapped from one to another via the mapping estimated by the cross-image flow module
        % For each tuple, the third image show the result 
        % where the landmarks at the second column are warped to the first image and 
        % the landmark results of the fourth image are warped from the first column. 
        % In this figure we compare the cross-image consistency of detected landmarks obtained by our method and a baseline that uses synthesized image pairs. We achieve this by transforming detected landmarks on one image to another image using the map estimated by the cross-image flow module. From the results, we can see landmarks detected by our method are significantly more consistent across different images.
        }
    \label{fig::consistency}
    \vspace{-10pt}
\end{figure}

Figure \ref{fig::consistency} shows the comparison between the detected landmarks obtained by our method and the baseline \cite{jakab2018unsupervised}.
Specifically, we transform the detected landmarks from one image to another 
and plot together the original detected landmarks as well as the transformed ones.
%·
Ideally, the original and the transformed landmarks should overlap each other seamlessly.
From Figure \ref{fig::consistency} it is clearly that landmarks detected by our method are more consistent across different images,
where the images are varying in various appearance factors such as gender.
Such results not only demonstrate the effectiveness of our method but also the necessity of learning from natural image pairs that contain both varying appearances and poses.
More results are available in the supplementary materials.

\begin{figure}
    \centering
    \subfloat[\emph{Disentanglement of Appearance and Landmark}. The synthesized results suggest that our encoders could disentangle the appearance/landmark-specific representations well ]{
        \includegraphics[width=1\textwidth]{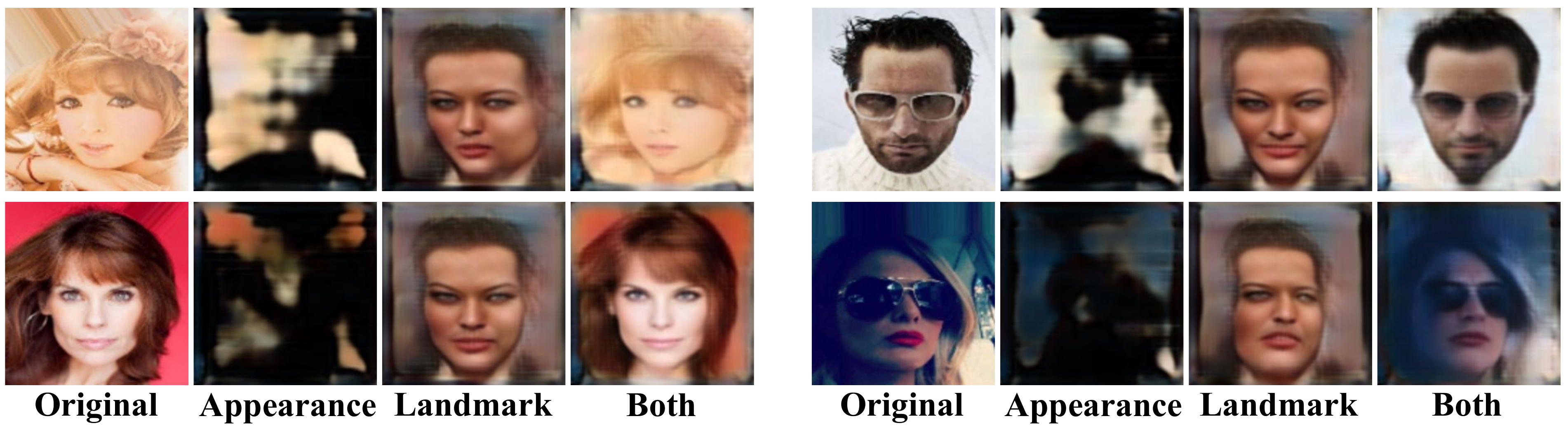}
        }

        \subfloat[\emph{Template Face}. When decoding with only the pose feature maps, the decoded images are consistent with the original images in terms of pose-related information,
        while share the same appearance-related information]{
        \includegraphics[width=1\textwidth]{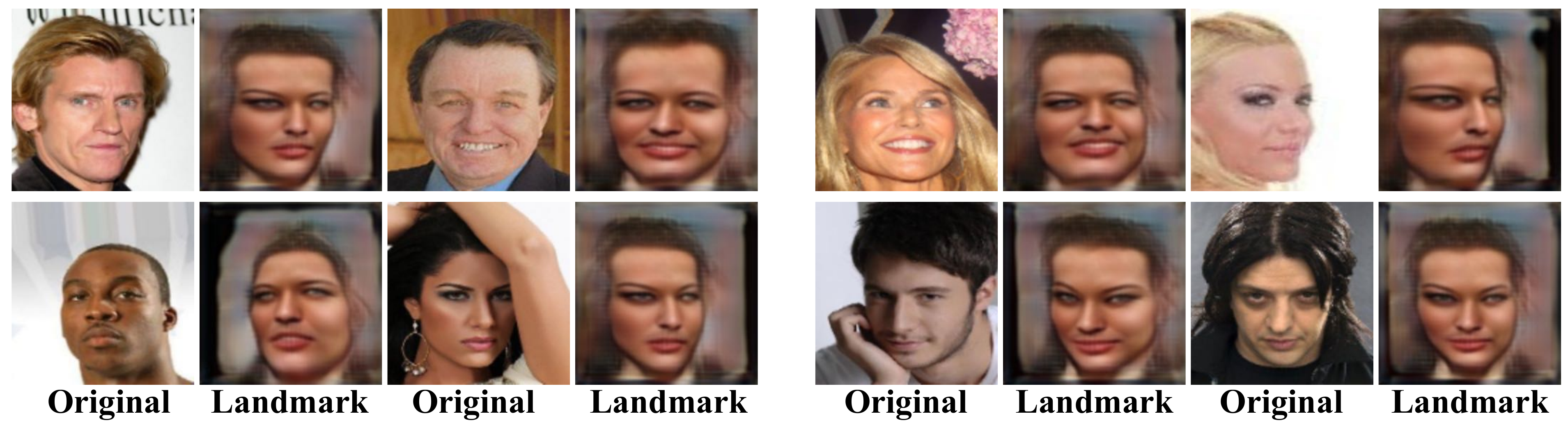}
        } 

    \caption{
        \textbf{Visualization and interpretation of our method} 
}
\label{fig::disentanglement}
    \vspace{-1.5em}
\end{figure}

\noindent{\textbf{Disentanglement of Appearance and Pose.}}
Another benefit of our framework is the disentanglement of appearance and pose for a given image.
%
%They tend to employ two encoders to represent them separately.
%
While it is directly related to the quality of detected landmarks,
the disentanglement is encouraged by our framework but not directly constrained. 
We thus investigate the completeness of disentanglement here.
Specifically, while the decoder takes both an appearance feature map and a pose feature map as the input,
we can replace one of the feature maps with a zero-valued map
and study the decoded image (referred as isolated image decoding).
%
%We refer to this study as isolated image decoding.

The results of isolated image decoding are included in Figure \ref{fig::disentanglement}.
When decoding with only the appearance feature maps,
it can be seen the decoded images contain only appearance information such as texture and color, 
without any pose-related structural information.
More importantly, when decoding with only the pose feature maps,
the decoded images are consistent with the original images in terms of pose-related information,
while sharing the same appearance-related information, which can be regarded as a \emph{templated face}.
These results demonstrate that the appearance and pose of the input image have been effectively disentangled by our method.
Moreover, they indicate that the decoder will at first transform the templated face into a specific pose according to the input pose,
then refine the transformed face according to the input appearance information.
Such a process is efficient and natural in the perspective of humans, 
which is surprisingly learned in a totally data-driven way.

\begin{wraptable}{r}{5cm}
   \vspace{-15pt}
\centering
    \caption{Semantic Alignment}
     \label{tab:se_align}
    \begin{tabular}{l c}
        \toprule 
        Method           & PCK @ 0.1 \\
        \midrule 
        PF-LOM \cite{ham2017proposal}  & 62.5 \\
        CNNGeo \cite{cnngeo}      & 71.9 \\
        A2Net  \cite{A2Net}  & 74.1 \\
        RTNs \cite{RTN}        & 76.2 \\
        WeakAlign \cite{weakalign}   & 75.8 \\
        NC-Net \cite{Neighbourhood}         & $78.9$ \\
        \midrule
        Ours             & \textbf{78.5} \\
        Ours w. landmark & \textbf{80.1}  \\
        \bottomrule
        \end{tabular}
        \vspace{-15pt}
\end{wraptable}

\subsection{Semantic Alignment} \label{exp:se_align}
Our cross-image flow module enhances the landmark consistency via the correspondences across different images, which is also the target of semantic alignment task.
We thus also verify the effectiveness of the cross-image flow module on the task of semantic alignment.
Specifically, semantic alignment aims to learn the dense correspondences of given image pairs which come from the same category but slightly differ in appearance.

\textbf{Setup.}
The standard evaluation protocol for semantic alignment is adopted and the dataset PF-PASCAL \cite{ham2017proposal} is divided into approximately 700 pairs for training, 300 pairs for validation and 300 pairs for testing. 
The performance is measured using the percentage of correct keypoints (PCK), which is number of correctly matched ground-truths.
%
%The model is trained on 8 GPUs with a total batch size of 128. Adam \misscite is adopted as our optimizer with learning rate of 0.0005.
%
In order to conduct an apple-to-apple comparison, we follow the training receipt of \cite{Neighbourhood}. More details are presented in the supplementary materials.

\textbf{Quantitative Evaluation.}
As shown in Table \ref{tab:se_align}, our method shows a competitive performance with the state-of-the-art methods.
When further equipped with the proposed cross-image cycle consistent framework, 
our method obtains a $1.2\%$ improvement over NC-NET \cite{Neighbourhood} on PF-PASCAL \cite{ham2017proposal}.
Moreover, in terms of efficiency, our method only requires 81 GFLOPs while NC-NET \cite{Neighbourhood} needs 471 GFLOPs.
The results in Table \ref{tab:se_align} clearly demonstrate the effectiveness of the proposed cross-image flow module,
which leads to higher performance and lower computational complexity.

\begin{figure}[t]
    \centering
    \includegraphics[width=1\textwidth]{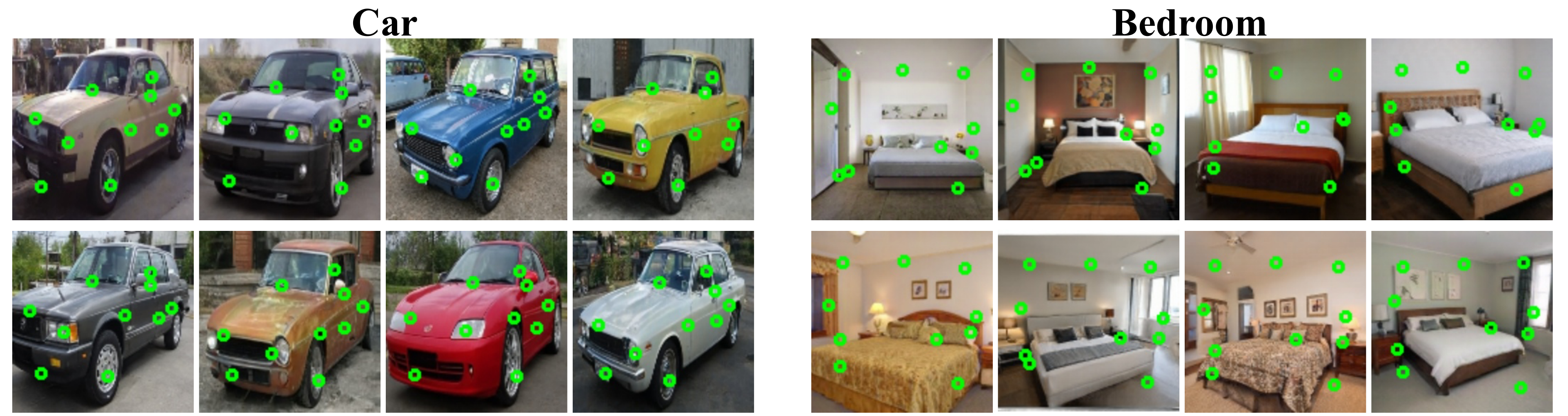}
    \caption{
        \textbf{Generalization.} The green dots denote the learned landmarks of our method
        }
    \label{fig::carandscene}
    \vspace{-15pt}
\end{figure}  

\subsection{Generalization to Car and Bedroom}
To show the generalization of our method, we apply the landmark discovery on car and bedroom. Training details and dataset statistics are available in our supplementary files.
Figure \ref{fig::carandscene} presents the detected landmarks.
In terms of a certain object-centric image collection \eg cars, our method could indentify a set of landmark points which still maintain the geometric and semantic consistency across instances. 
When meeting scenes which contain not only multiple objects but also include the relationship among them, the learned landmarks pour more attention on the geometric structural information \ie the layout lines and orientations are encoded.
\section{Conclusion}\label{sec:conclusion}

In this paper, we propose a cross-image cycle consistency framework to enable the unsupervised landmark learning on unaligned image pairs sampled from a collection of natural images.
Compared to synthesized image pairs used by previous methods, the unaligned pairs can vary in both appearance and pose.
Such unaligned pairs  can encourage the proposed framework to better disentangle the appearance and landmark, leading to a clear improvement of landmarks.
Moreover, a cross-image flow module further enhances the cross-image consistency of detected landmarks.
On landmark detection and a closely-related task semantic alignment, extensive experiments validate that our approach outperforms existing methods by a large margin.
Furthermore, the visualization and interpretation of our approach demonstrate that the models learned by our framework align well with the motivation of unsupervised landmark detection.

\section*{Broader Impacts}
Landmarks adaptively distribute a set of points to semantical and geometrical positions on an object, providing informative cues for object analysis.
Our framework learns to detect landmarks in an unsupervised manner from unpaired data, which are massive online.
Specifically, millions of unlabelled images are shared on the internet daily.
Our proposed approach requires no human annotations and can work well with in-the-wild image collections.
In addition, the correspondence of the landmark across different instances can be exploited as the glue that links disparate visual percepts for the fundamental vision task from optical flow and tracking to action recognition and 3D reconstruction.
Another impressive thing is that the template face emerged in our work, which shows the potential for designs of generative models and network defense due to the preservation of object structure without appearance information such as texture and color.
Because our work focuses on the structure representation, it does not present any foreseeable ethical aspects and societal consequences.

{\small
\bibliographystyle{abbrvnat}
\bibliography{references}

\begin{thebibliography}{39}
\providecommand{\natexlab}[1]{#1}
\providecommand{\url}[1]{\texttt{#1}}
\expandafter\ifx\csname urlstyle\endcsname\relax
  \providecommand{\doi}[1]{doi: #1}\else
  \providecommand{\doi}{doi: \begingroup \urlstyle{rm}\Url}\fi

\bibitem[Burgos-Artizzu et~al.(2013)Burgos-Artizzu, Perona, and
  Doll{\'a}r]{RCPR}
X.~P. Burgos-Artizzu, P.~Perona, and P.~Doll{\'a}r.
\newblock Robust face landmark estimation under occlusion.
\newblock In \emph{Proc. ICCV}, 2013.

\bibitem[Cao et~al.(2017)Cao, Simon, Wei, and Sheikh]{openpose}
Z.~Cao, T.~Simon, S.-E. Wei, and Y.~Sheikh.
\newblock Realtime multi-person 2d pose estimation using part affinity fields.
\newblock In \emph{Proc. CVPR}, 2017.

\bibitem[Chen et~al.(2016)Chen, Duan, Houthooft, Schulman, Sutskever, and
  Abbeel]{infogan}
X.~Chen, Y.~Duan, R.~Houthooft, J.~Schulman, I.~Sutskever, and P.~Abbeel.
\newblock Infogan: Interpretable representation learning by information
  maximizing generative adversarial nets.
\newblock In \emph{Advances in Neural Information Processing Systems}, 2016.

\bibitem[Deng et~al.(2009)Deng, Dong, Socher, Li, Li, and Fei-Fei]{imagenet}
J.~Deng, W.~Dong, R.~Socher, L.-J. Li, K.~Li, and L.~Fei-Fei.
\newblock Imagenet: A large-scale hierarchical image database.
\newblock In \emph{Proc. CVPR}, 2009.

\bibitem[Denton et~al.(2017)]{denton2017unsupervised}
E.~L. Denton et~al.
\newblock Unsupervised learning of disentangled representations from video.
\newblock In \emph{Advances in Neural Information Processing Systems}, 2017.

\bibitem[Desjardins et~al.(2012)Desjardins, Courville, and
  Bengio]{desjardins2012disentangling}
G.~Desjardins, A.~Courville, and Y.~Bengio.
\newblock Disentangling factors of variation via generative entangling.
\newblock \emph{arXiv preprint arXiv:1210.5474}, 2012.

\bibitem[Du et~al.(2015)Du, Wang, and Wang]{du2015hierarchical}
Y.~Du, W.~Wang, and L.~Wang.
\newblock Hierarchical recurrent neural network for skeleton based action
  recognition.
\newblock In \emph{Proc. CVPR}, 2015.

\bibitem[Dundar et~al.(2020{\natexlab{a}})Dundar, Shih, Garg, Pottorf, Tao, and
  Catanzaro]{abiv}
A.~Dundar, K.~J. Shih, A.~Garg, R.~Pottorf, A.~Tao, and B.~Catanzaro.
\newblock Unsupervised disentanglement of pose, appearance and background from
  images and videos.
\newblock \emph{arXiv preprint arXiv:2001.09518}, 2020{\natexlab{a}}.

\bibitem[Dundar et~al.(2020{\natexlab{b}})Dundar, Shih, Garg, Pottorf, Tao, and
  Catanzaro]{video_inter}
A.~Dundar, K.~J. Shih, A.~Garg, R.~Pottorf, A.~Tao, and B.~Catanzaro.
\newblock Unsupervised disentanglement of pose, appearance and background from
  images and videos.
\newblock \emph{arXiv preprint arXiv:2001.09518}, 2020{\natexlab{b}}.

\bibitem[Everingham et~al.(2010)Everingham, Van~Gool, Williams, Winn, and
  Zisserman]{pascal}
M.~Everingham, L.~Van~Gool, C.~K. Williams, J.~Winn, and A.~Zisserman.
\newblock The pascal visual object classes (voc) challenge.
\newblock \emph{International Journal of Computer Vision}, 2010.

\bibitem[Ham et~al.(2017)Ham, Cho, Schmid, and Ponce]{ham2017proposal}
B.~Ham, M.~Cho, C.~Schmid, and J.~Ponce.
\newblock Proposal flow: Semantic correspondences from object proposals.
\newblock \emph{IEEE Trans. on Pattern Analysis and Machine Intelligence},
  2017.

\bibitem[He et~al.(2016)He, Zhang, Ren, and Sun]{resnet}
K.~He, X.~Zhang, S.~Ren, and J.~Sun.
\newblock Deep residual learning for image recognition.
\newblock In \emph{Proc. CVPR}, 2016.

\bibitem[Hinton and Salakhutdinov(2006)]{hintonbz}
G.~E. Hinton and R.~R. Salakhutdinov.
\newblock Reducing the dimensionality of data with neural networks.
\newblock \emph{Science}, 2006.

\bibitem[Jakab et~al.(2018)Jakab, Gupta, Bilen, and
  Vedaldi]{jakab2018unsupervised}
T.~Jakab, A.~Gupta, H.~Bilen, and A.~Vedaldi.
\newblock Unsupervised learning of object landmarks through conditional image
  generation.
\newblock In \emph{Advances in Neural Information Processing Systems}, 2018.

\bibitem[Jeon et~al.(2019)Jeon, Min, Kim, and Sohn]{jeon2019joint}
S.~Jeon, D.~Min, S.~Kim, and K.~Sohn.
\newblock Joint learning of semantic alignment and object landmark detection.
\newblock In \emph{Proc. ICCV}, 2019.

\bibitem[Johnson et~al.(2016)Johnson, Alahi, and Fei-Fei]{perceptual}
J.~Johnson, A.~Alahi, and L.~Fei-Fei.
\newblock Perceptual losses for real-time style transfer and super-resolution.
\newblock In \emph{Proc. ECCV}, 2016.

\bibitem[Kim et~al.(2018)Kim, Lin, JEON, Min, and Sohn]{RTN}
S.~Kim, S.~Lin, S.~R. JEON, D.~Min, and K.~Sohn.
\newblock Recurrent transformer networks for semantic correspondence.
\newblock In \emph{Advances in Neural Information Processing Systems}, 2018.

\bibitem[Kingma and Ba(2014)]{adam}
D.~P. Kingma and J.~Ba.
\newblock Adam: A method for stochastic optimization.
\newblock \emph{arXiv preprint arXiv:1412.6980}, 2014.

\bibitem[Koestinger et~al.(2011)Koestinger, Wohlhart, Roth, and Bischof]{AFLW}
M.~Koestinger, P.~Wohlhart, P.~M. Roth, and H.~Bischof.
\newblock Annotated facial landmarks in the wild: A large-scale, real-world
  database for facial landmark localization.
\newblock 2011.

\bibitem[Lee et~al.(2018)Lee, Tseng, Huang, Singh, and Yang]{DRIT}
H.-Y. Lee, H.-Y. Tseng, J.-B. Huang, M.~K. Singh, and M.-H. Yang.
\newblock Diverse image-to-image translation via disentangled representations.
\newblock In \emph{Proc. ECCV}, 2018.

\bibitem[Liu et~al.(2015)Liu, Luo, Wang, and Tang]{celeba}
Z.~Liu, P.~Luo, X.~Wang, and X.~Tang.
\newblock Deep learning face attributes in the wild.
\newblock In \emph{Proc. ICCV}, 2015.

\bibitem[Lorenz et~al.(2019)Lorenz, Bereska, Milbich, and
  Ommer]{lorenz2019unsupervised}
D.~Lorenz, L.~Bereska, T.~Milbich, and B.~Ommer.
\newblock Unsupervised part-based disentangling of object shape and appearance.
\newblock In \emph{Proc. CVPR}, 2019.

\bibitem[Minderer et~al.(2019)Minderer, Sun, Villegas, Cole, Murphy, and
  Lee]{video_dynamic}
M.~Minderer, C.~Sun, R.~Villegas, F.~Cole, K.~P. Murphy, and H.~Lee.
\newblock Unsupervised learning of object structure and dynamics from videos.
\newblock In \emph{Advances in Neural Information Processing Systems}, 2019.

\bibitem[Nair and Hinton(2010)]{relu}
V.~Nair and G.~E. Hinton.
\newblock Rectified linear units improve restricted boltzmann machines.
\newblock In \emph{International Conference on Machine Learning}, 2010.

\bibitem[Newell et~al.(2016)Newell, Yang, and Deng]{hourglass}
A.~Newell, K.~Yang, and J.~Deng.
\newblock Stacked hourglass networks for human pose estimation.
\newblock In \emph{Proc. ECCV}, 2016.

\bibitem[Rocco et~al.(2017)Rocco, Arandjelovic, and Sivic]{cnngeo}
I.~Rocco, R.~Arandjelovic, and J.~Sivic.
\newblock Convolutional neural network architecture for geometric matching.
\newblock In \emph{Proc. CVPR}, 2017.

\bibitem[Rocco et~al.(2018{\natexlab{a}})Rocco, Arandjelovi{\'c}, and
  Sivic]{weakalign}
I.~Rocco, R.~Arandjelovi{\'c}, and J.~Sivic.
\newblock End-to-end weakly-supervised semantic alignment.
\newblock In \emph{Proc. CVPR}, 2018{\natexlab{a}}.

\bibitem[Rocco et~al.(2018{\natexlab{b}})Rocco, Cimpoi, Arandjelovi{\'c},
  Torii, Pajdla, and Sivic]{Neighbourhood}
I.~Rocco, M.~Cimpoi, R.~Arandjelovi{\'c}, A.~Torii, T.~Pajdla, and J.~Sivic.
\newblock Neighbourhood consensus networks.
\newblock In \emph{Advances in Neural Information Processing Systems},
  2018{\natexlab{b}}.

\bibitem[Seo et~al.(2018)Seo, Lee, Jung, Han, and Cho]{A2Net}
P.~H. Seo, J.~Lee, D.~Jung, B.~Han, and M.~Cho.
\newblock Attentive semantic alignment with offset-aware correlation kernels.
\newblock In \emph{Proc. ECCV}, 2018.

\bibitem[Simonyan and Zisserman(2015)]{vgg}
K.~Simonyan and A.~Zisserman.
\newblock Very deep convolutional networks for large-scale image recognition.
\newblock In \emph{International Conference on Learning Representations}, 2015.

\bibitem[Sun et~al.(2013)Sun, Wang, and Tang]{cascadedCNN}
Y.~Sun, X.~Wang, and X.~Tang.
\newblock Deep convolutional network cascade for facial point detection.
\newblock In \emph{Proc. CVPR}, 2013.

\bibitem[Thewlis et~al.(2017{\natexlab{a}})Thewlis, Bilen, and
  Vedaldi]{landmarkfactorize}
J.~Thewlis, H.~Bilen, and A.~Vedaldi.
\newblock Unsupervised learning of object landmarks by factorized spatial
  embeddings.
\newblock In \emph{Proc. ICCV}, 2017{\natexlab{a}}.

\bibitem[Thewlis et~al.(2017{\natexlab{b}})Thewlis, Bilen, and
  Vedaldi]{thewlis2017unsupervised}
J.~Thewlis, H.~Bilen, and A.~Vedaldi.
\newblock Unsupervised learning of object frames by dense equivariant image
  labelling.
\newblock In \emph{Advances in Neural Information Processing Systems},
  2017{\natexlab{b}}.

\bibitem[Thewlis et~al.(2019)Thewlis, Albanie, Bilen, and
  Vedaldi]{thewlis2019unsupervised}
J.~Thewlis, S.~Albanie, H.~Bilen, and A.~Vedaldi.
\newblock Unsupervised learning of landmarks by descriptor vector exchange.
\newblock In \emph{Proc. ICCV}, 2019.

\bibitem[Yan et~al.(2018)Yan, Xiong, and Lin]{yan2018spatial}
S.~Yan, Y.~Xiong, and D.~Lin.
\newblock Spatial temporal graph convolutional networks for skeleton-based
  action recognition.
\newblock In \emph{Thirty-second AAAI conference on artificial intelligence},
  2018.

\bibitem[Zhang et~al.(2014{\natexlab{a}})Zhang, Shan, Kan, and Chen]{CFAN}
J.~Zhang, S.~Shan, M.~Kan, and X.~Chen.
\newblock Coarse-to-fine auto-encoder networks (cfan) for real-time face
  alignment.
\newblock In \emph{Proc. ECCV}, 2014{\natexlab{a}}.

\bibitem[Zhang et~al.(2018)Zhang, Guo, Jin, Luo, He, and Lee]{Zhang_2018_CVPR}
Y.~Zhang, Y.~Guo, Y.~Jin, Y.~Luo, Z.~He, and H.~Lee.
\newblock Unsupervised discovery of object landmarks as structural
  representations.
\newblock In \emph{Proc. CVPR}, 2018.

\bibitem[Zhang et~al.(2014{\natexlab{b}})Zhang, Luo, Loy, and Tang]{MTCNN}
Z.~Zhang, P.~Luo, C.~C. Loy, and X.~Tang.
\newblock Facial landmark detection by deep multi-task learning.
\newblock In \emph{Proc. ECCV}, 2014{\natexlab{b}}.

\bibitem[Zhang et~al.(2015)Zhang, Luo, Loy, and Tang]{TCDCN}
Z.~Zhang, P.~Luo, C.~C. Loy, and X.~Tang.
\newblock Learning deep representation for face alignment with auxiliary
  attributes.
\newblock \emph{IEEE Trans. on Pattern Analysis and Machine Intelligence},
  2015.

\end{thebibliography}
}
\section*{Appendix}\label{sec:intro}
\textbf{Architectures.}
The decoder $E$ comprises of sequential blocks with two convolutional layers each. 
The input to each successive block, except the first one, is upsampled two times through bilinear interpolation,
while the number of feature channels is halved.
The first block starts with 256 channels, and a minimum of 32 channels are maintained till a tensor with the same spatial dimensions as the input image is obtained. 
A final convolutional layer regresses the three RGB channels without non-linear activation.

\textbf{Training Protocol for Facial Landmark Detection.} 
We provide more details for training on facial landmark.
We first pretrain the $E_a$, $E_p$ and $D$ utilizing paired data following \cite{jakab2018unsupervised} and semantic flow module following \cite{Neighbourhood}.
We stop gradients for the first swapping reconstruction image $a_x$, $a_y$ to prevent the optimization collapse into a trivial solution.
The models are trained on $4$ GPUs with $50$ images per GPU for $200$ epochs. 
All the weights are initialized with random Gaussian distribution ($std=0.01$), optimized by Adam.
The initial learning rate is $5\times 10^{-4}$ and then divided by a factor of $10$ at $100$ and $150$ epoch respectively. 
The weight decay is set to $10^{-4}$. 

\begin{figure}[h]
  \centering
  \includegraphics[width=1\textwidth]{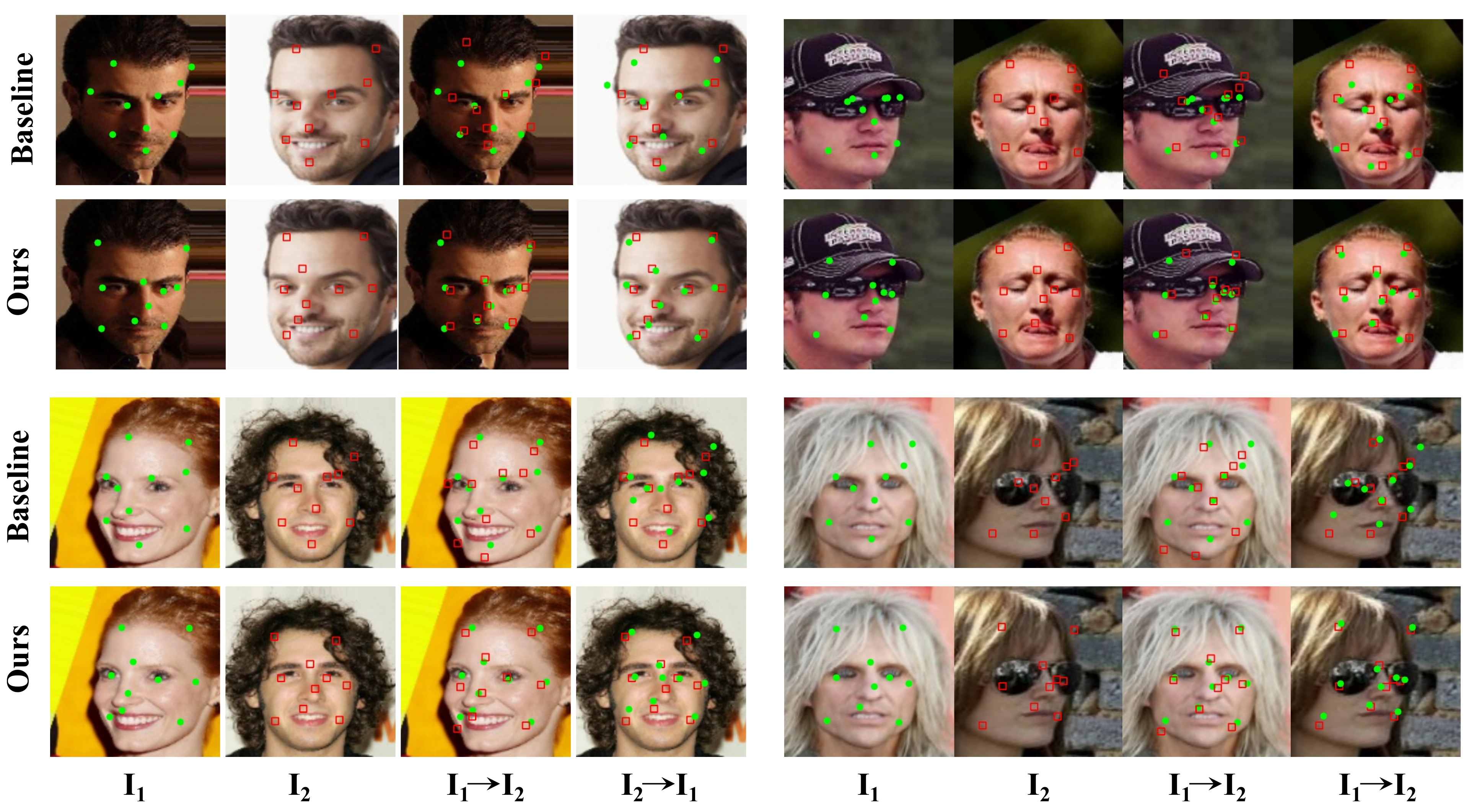}
  \caption{More results on Cross-Image Landmark Consistency}
  \label{fig:consis}
\end{figure}

\textbf{More Results.}
In Fig \ref{fig:consis}, we show more results on Cross-Image Landmark Consistency. Obviously, learning from different instances (Ours) allows to obtain more consistent landmarks than those from the same (Baseline) \cite{jakab2018unsupervised}.

\textbf{Training Protocol for Semantic Alignment.}
For semantic alignment, the model is firstly trained for 5 epochs using Adam, 
with a learning rate of $5\times 10^{-4}$ and keeping the feature extraction layer weights fixed.
And then the model is subsequently finetuned for 5 more epochs, training both the feature extraction and the semantic flow module , with a learning rate of $1\times 10^{-5}$.

\textbf{Dataset statistics for Car and Bedroom}
For car and bedroom landmark detection, we collected data both from LSUN dataset and train the landmark detection with the same setting for face landmark detection.

\end{document}